\documentclass[runningheads]{llncs}

\usepackage{eccv}

\usepackage{eccvabbrv}

\usepackage{graphicx}
\usepackage{booktabs}

\usepackage[accsupp]{axessibility}  

\usepackage{tikz}
\usepackage{rotating}
\newcommand{\spheading}[2][6.4em]{
  \rotatebox{90}{\parbox{#1}{\raggedright #2}}}
\usepackage{multirow,tabularx}
\usepackage{pifont} 
\usepackage{graphicx}
\usepackage{color,xcolor, colortbl}
\definecolor{Gray2}{gray}{0.95}
\definecolor{blue1}{RGB}{68,114,196}
\definecolor{blue2}{RGB}{198,212,238}
\definecolor{orange1}{RGB}{252,163,19}
\definecolor{orange2}{RGB}{252,219,180}
\definecolor{red1}{RGB}{192,0,0}
\definecolor{red2}{RGB}{237,179,178}
\usepackage{adjustbox}
\usepackage{pifont} 
\usepackage{overpic}
\usepackage{subfloat}
\usepackage{enumitem}

\usepackage{makecell}
\usepackage{array}
\usepackage{caption}
\usepackage{wrapfig}
\usepackage{subcaption} %

\newsavebox{\bluecircle}
\begin{lrbox}{\bluecircle}
    \raisebox{-0.1ex}{\tikz\filldraw[fill=blue2, draw=blue1] (0,0) circle [radius=0.1cm];}
\end{lrbox}

\newsavebox{\orangecircle}
\begin{lrbox}{\orangecircle}
    \raisebox{-0.1ex}{\tikz\filldraw[fill=orange2, draw=orange1] (0,0) circle [radius=0.1cm];}
\end{lrbox}

\newsavebox{\redcircle}
\begin{lrbox}{\redcircle}
    \raisebox{-0.1ex}{\tikz\filldraw[fill=red2, draw=red1] (0,0) circle [radius=0.1cm];}
\end{lrbox}

\newcommand\blfootnote[1]{%
\begingroup
\renewcommand\thefootnote{}\footnote{#1}%
\addtocounter{footnote}{-1}%
\endgroup
}

\usepackage{wasysym}

\usepackage{orcidlink}
\usepackage{marvosym}
\begin{document}

\title{OphNet: A Large-Scale Video Benchmark for Ophthalmic Surgical Workflow Understanding} 

\titlerunning{OphNet: A Benchmark for Ophthalmic Surgical Workflow Understanding}

\authorrunning{M. Hu et al.}

\author{
Ming Hu\inst{1,2,3}$^{*}$ \and
Peng Xia\inst{1,3}$^{*}$ \and
Lin Wang\inst{5}$^{*}$ \and
Siyuan Yan\inst{1,2} \and
Feilong Tang\inst{1,3} \\
Zhongxing Xu\inst{7} \and
Yimin Luo\inst{6} \and 
Kaimin Song\inst{3} \and
Jurgen Leitner\inst{2} \\
Xuelian Cheng\inst{1} \and
Jun Cheng\inst{8} \and
Chi Liu\inst{9} \and
Kaijing Zhou\inst{4}$^{\dagger}$\and
Zongyuan Ge\inst{1,2,3}$^{\dagger}$
}

\institute{
$^1$AIM Lab, Faculty of IT, $^2$Faculty of Engineering, Monash University \\
$^3$Airdoc-Monash Research, Airdoc \quad
$^4$Eye Hospital, Wenzhou Medical University \\
$^5$Bosch Corporate Research \quad
$^6$King's College London \quad
$^7$Cornell University \\
$^8$Institute for Infocomm Research, A*STAR \\
$^9$Faculty of Data Science, City University of Macau
}

\maketitle

\begin{abstract}
Surgical scene perception via videos is critical for advancing robotic surgery, telesurgery, and AI-assisted surgery, particularly in ophthalmology. However, the scarcity of diverse and richly annotated video datasets has hindered the development of intelligent systems for surgical workflow analysis. Existing datasets face challenges such as small scale, lack of diversity in surgery and phase categories, and absence of time-localized annotations. These limitations impede action understanding and model generalization validation in complex and diverse real-world surgical scenarios. To address this gap, we introduce OphNet, a large-scale, expert-annotated video benchmark for ophthalmic surgical workflow understanding. OphNet features: 1) A diverse collection of 2,278 surgical videos spanning 66 types of cataract, glaucoma, and corneal surgeries, with detailed annotations for 102 unique surgical phases and 150 fine-grained operations. 2) Sequential and hierarchical annotations for each surgery, phase, and operation, enabling comprehensive understanding and improved interpretability. 3) Time-localized annotations, facilitating temporal localization and prediction tasks within surgical workflows. With approximately 285 hours of surgical videos, OphNet is about 20 times larger than the largest existing surgical workflow analysis benchmark. Code and dataset are available at: \url{https://minghu0830.github.io/OphNet-benchmark/}.

\end{abstract}

\blfootnote{$*$ Equal contribution ~~~~~ $\dagger$ Corresponding author}

\section{Introduction}
\label{sec:intro}
As surgical robot platforms such as the da Vinci\textsuperscript\textregistered~surgical system become increasingly sophisticated, there is growing interest in integrating enhanced intelligence into scenarios like minimally invasive surgery~\cite{Forslund_Jacobsen2020-gc, s23208503, 10160746}. The advancements in machine vision perception empower these robotic systems to autonomously recognize and adapt to the intricacies of surgical environments, without relying on binary instrument usage signals, RFID tags, sensor data from tracking devices, or other signal information that necessitates laborious manual annotations or additional equipment installations~\cite{Jin2018-ek}. This autonomy includes the capacity to identify anatomical structures, detect anomalies, and adjust surgical plans in real-time, which is crucial in dynamic and unpredictable surgical settings. In recent years, especially in endoscopy and ophthalmic surgery, the application of deep learning has demonstrated considerable promise in bolstering these autonomous capabilities. This encompasses the analysis of surgical workflows~\cite{Jin2018-ek, Yu2019, Bar2020}, segmentation of instruments and anatomy~\cite{CATARACTS,grammatikopoulou2022cadis, ni2019raunet}, and depth estimation~\cite{zha2023endosurf}, among others.

Automatic video surgical workflow understanding is a fundamental yet challenging problem for developing computer-assisted and robotic-assisted surgery, which can be divided into internal (e.g., laparoscopic and endoscopic~\cite{Jin2018-ek, Pan2023-je, Shi2020-fb}) and external (e.g., operating room and nursing procedure~\cite{ming2023nurvid}) analysis. In addition to promoting the development of intelligent surgery, it also greatly benefits surgical documentation, education, and training~\cite{Twinanda2016EndoNetAD, Czempiel2020, czempiel2021opera}. Baret et al.~\cite{Bar2020} showed networks, like Inflated 3D ConvNet (I3D)~\cite{i3d} that utilize spatiotemporal convolutions, require a relatively extensive dataset for effective training. In their study, the model achieves an accuracy exceeding 80\% when trained on 100 videos, with a progressive improvement as the sample size surpasses 700. However, the highly efficient and rapidly evolving deep learning technologies for surgical workflow analysis are currently limited by the following shortcomings in current video benchmarks: 
 \textbf{1) Small-scale:} the majority of video datasets contain no more than 100 videos. For example, the CATARACTS~\cite{DBLP:conf/icpr/GhamsarianTPSS20} and CatRelDet~\cite{DBLP:conf/icpr/GhamsarianTPSS20} datasets contain only 50 and 21 surgical videos, respectively. These datasets are relatively small, insufficient for large-scale validation.
 \textbf{2) Limited categories of surgeries and phases:} almost all ophthalmic surgical video datasets only include cataract surgery and do not further classify specific types of surgeries. Additionally, the number of phase categories is also limited, like CatRelDet~\cite{DBLP:conf/icpr/GhamsarianTPSS20} only contains 4 different phase labels, which is insufficient to meet the requirements for evaluation in real clinical environments.
\textbf{3) Coarse-grained annotation:} due to annotation costs, existing benchmarks often have coarse-grained action definitions. For example, adhesive injection may occur in two different phases: main incision and capsulorhexis, so it may be classified into different phase categories. Coarse-grained action definitions may lead to annotation bias. 
\textbf{4) Single time-boundary annotation:} they only annotate designated phases in the videos, ignoring the continuity across different stages of ophthalmic surgery, as well as the hierarchical relationship between surgery, phase, and operation. Simpler datasets, such as LensID~\cite{LensID}, are limited to binary classification tasks distinguishing lens implantation from other irrelevant phases.
\textbf{5) Uniform domain:} the videos are meticulously collected, and while this ensures video quality, the uniform style is not conducive to testing the model's domain generalization ability.

\begin{table*}[t!]
\tiny
\begin{center}
\resizebox{1\linewidth}{!}{
\begin{tabular}{p{0.09\linewidth}|p{0.20\linewidth}|p{0.05\linewidth}|p{0.05\linewidth}|p{0.03\linewidth}|p{0.04\linewidth}|p{0.06\linewidth}| 
p{0.05\linewidth}|p{0.03\linewidth}|p{0.03\linewidth}|p{0.03\linewidth}}
\toprule
\multirow{5}{*}{\centerline{\spheading{Protocol}}} &\multicolumn{6}{c|}{Dataset Properties} & \multicolumn{4}{c}{Tasks} \\ 
&\cline{1-10}
& \spheading{Datasets}  
& \spheading{No. of Videos} 
& \spheading{No. of Action Segments} 
& \spheading{No. of Surgery Categories}
& \spheading{No. of Action Categories}  
& \spheading{Total Duration} 
& \spheading{Multi-Surgery Presence Recognition} 
& \spheading{Phase Recognition} 
& \spheading{Phase Localization} 
& \spheading{Phase Prediction} \\
\midrule
& Cholec120~\cite{nwoye2022data} & 120 & - & 1 & 7 & 76.2h & \ding{55} & \ding{51} & \ding{55} & \ding{55}\\

& SurgicalActions160~\cite{DBLP:journals/mta/SchoeffmannHKPM18} & 160 & 160 & 1& 16 & 0.2h & \ding{55} & \ding{51} & \ding{55} & \ding{55}\\

 & HeiCo~\cite{maier2021heidelberg, ross2021comparative} & 30 & - & 3 & 14 & 2.8h & \ding{55} & \ding{51} & \ding{51} & \ding{51}  \\

Endo\&Lap & EndoVis 2021~\cite{wagner2023comparative}  & 33 & 250 & 1 & 7 & 22.0h & \ding{55} & \ding{51} & \ding{55} & \ding{55}  \\

& PitVis~\cite{PitVis}  & 25 & 287 & 1 & 17 & 33.3h & \ding{55} & \ding{51} & \ding{55} & \ding{55}\\

& CholecT50~\cite{nwoye2022rendezvous} & 50 & - & 1 & 10 & 44.7h & \ding{55} & \ding{51} & \ding{51} & \ding{51} \\

& AutoLaparo~\cite{wang2022autolaparo} & 21 & 300 & 1 & 7 & 23.1h & \ding{55} & \ding{51} & \ding{51} & \ding{51} \\

\midrule
& LensID~\cite{LensID} & 100 & 2,440 & 1 & 2 & 11.7h & \ding{55} & \ding{51} & \ding{55} & \ding{55}\\

& Cataract-101~\cite{DBLP:conf/mmsys/SchoeffmannTSMP18}  & 101 & 1,266 & 1 & 10  & 14.0h & \ding{55} & \ding{51} & \ding{51} & \ding{51} \\
OphScope & CatRelDet~\cite{DBLP:conf/icpr/GhamsarianTPSS20}  & 21 & 2,400 & 1 & 4 & 2.0h & \ding{55} & \ding{51} & \ding{55} & \ding{55}\\

& CATARACTS~\cite{CATARACTS} & 50 & 1,536 & 1 & 19 & 20.0h & \ding{55} & \ding{51} & \ding{51} & \ding{51} \\

& Cataract-1K~\cite{ghamsarian2023cataract} & 1,000 & 931 & 1 & 12 & 118.7h & \ding{55} & \ding{51} & \ding{51} & \ding{51} \\ 

\rowcolor{Gray2}
& \textbf{OphNet(Ours)}  & \textbf{2,278} & \textbf{9,795} & \textbf{66} & \textbf{150}  & \textbf{284.8h} & \ding{51} & \ding{51} & \ding{51} & \ding{51}  \\
\bottomrule
\end{tabular}}
\end{center}
\caption{The statistics comparison among existing workflow analysis datasets and our OphNet. Compared to other datasets, OphNet focuses on more comprehensive coverage of various surgery, phase and operation categories, collects a large number of videos, totaling 284.8 hours, and also enables a variety of recognition, localization and prediction tasks. OphNet demonstrates considerable competitiveness in both its scale and the richness of its labels. For instance, Cholec120~\cite{nwoye2022data}, Cholec80~\cite{Twinanda2016EndoNetAD}, m2cai-workflow and LapChole~\cite{stauder2016tum} form one series, whereas CholecT50~\cite{nwoye2022rendezvous}, CholecT45~\cite{nwoye2022rendezvous}, and CholecT40~\cite{nwoye2020recognition} comprise another series. We have excluded the following scenarios from our comparison: (1) non-open-source datasets such as Bypass170~\cite{8509608}, ESD~\cite{10.1007/978-3-031-43996-4_47}, Yu's~\cite{Yu2019}, etc.; (2) a superset of multiple open-source or non-open-source datasets, like Cholec207~\cite{Bar2020}, etc.; (3) datasets employed for lesion, anatomy, and instrument classification and segmentation, such as SUN-SEG~\cite{Ji2022}, CVC-ClinicDB~\cite{BERNAL201599}, ROBUST-MIS~\cite{RO2021101920}, Mesejo's~\cite{7442848}, Cata7~\cite{ni2019raunet}, etc., anomaly detection such as PolypDiag~\cite{tian2022contrastive} (from Hyper-Kvasir~\cite{Borgli2020} and LDPolypVideo~\cite{Yiting2021}), Kvasir-Capsule~\cite{Smedsrud2021}, etc., and other datasets not dedicated to workflow analysis. It's worth mentioning that even in comparison with the above datasets, OphNet demonstrates considerable competitiveness in both its scale and the richness of its labels. \emph{Endo\&Lap} denotes the endoscopic and laparoscopic protocol, \emph{OphScope} denotes the ophthalmic microscope protocol. We choose the latest version for comparison in cases where datasets have multiple supplementary updates.}
\label{dataset_comparison}
\end{table*}

While some works have explored semi-supervised and self-supervised learning strategies~\cite{bodenstedt2017unsupervised, ross2018exploiting, yengera2018less, yu2018learning} to alleviate the cost of annotations or use only a small fraction of available labels, these approaches still lack competitiveness in performance compared to fully supervised learning. This deficiency in performance is impeding the widespread clinical application of these strategies. To address the shortage of sufficient labeled datasets, we construct OphNet, a large-scale and expert-level video benchmark with high diversity, for ophthalmic surgical workflow understanding. The main advantages of OphNet are as follows:

\begin{itemize}[leftmargin=*]
\item[$\bullet$]  \textbf{Largest scale and diversity:}
to the best of our knowledge, OphNet is currently the largest and most richly labeled dataset for surgical workflow analysis. It contains a number of videos 20 times greater than the current largest benchmark in ophthalmic surgery and far exceeds datasets in more established fields such as endoscopy. Additionally, OphNet includes the greatest variety of different types of surgeries, encompassing 66 different surgeries such as cataract, glaucoma, and corneal surgeries, along with 102 unique surgical phases and 150 distinct operations. This diversity significantly surpasses that of previous research.

\item[$\bullet$]  \textbf{Fine-grained, sequential and hierarchical annotation:} 
we have meticulously selected a subset of videos for annotation localization, with each video being annotated for an average of 22 operations. Additionally, we provide exquisite annotations at the levels of surgery, phase, and operation, catering to the requirements for training specific challenge models. This annotation design aims to offer a multifaceted understanding of surgical protocols, accommodate the nuances of each distinct surgery, and enhance the usability and interpretability of our dataset.

\item[$\bullet$] \textbf{Expert-level manual annotation:} 
the annotation work for OphNet was completed by ten experienced ophthalmologists and five individuals with ophthalmic experience, encompassing, but not limited to, standardization of definitions for surgery, phase, and operation labels, video filtering, classification and localization annotations, and secondary verification, among others. Expert-level annotators ensure the quality and professionalism of OphNet.
\end{itemize}

\section{Related Work}
\label{sec:related_work1}
\noindent \textbf{Surgical Workflow Understanding.} Beyond its therapeutic advantages, minimally invasive surgery also provides the capability for operative video recording. These videos can be stored and later utilized for various purposes such as cognitive training, skill assessment, and surgical workflow analysis~\cite{yuan2024hecvl, yuan2023learning}. Techniques derived from the broader field of video content analysis and representation are increasingly being incorporated into the surgical realm~\cite{Loukas2018, Yu2019, Bar2020}. A typical surgical workflow can be defined by a sequence of tasks or events, including patient positioning, incision, dissection, and suturing. These events are influenced not only by the specific type of surgery but also by the individual surgeon's proficiency and technique. Consequently, a comprehensive understanding of the surgical workflow necessitates a thorough examination of the temporal, spatial, and contextual facets of these tasks.

\noindent \textbf{Weakly-Supervised Video Learning.} 
\label{sec:related_work2}
Substantial pioneering work has been undertaken in the realms of video understanding~\cite{wu2023revisiting, bike, lin2019bmn,SSTAP, zhang2022actionformer, alwassel2021tsp, sato2023prompt, s23115024}. Since even a small amount of videos easily comprises several million frames, methods that do not rely on a frame-level annotation are of special importance. Weakly-supervised video learning makes use of loosely labeled data to train models, thereby obviating the necessity for exhaustively annotated training data. These weak labels might manifest in a variety of forms, including video-level labels or partial labels, which are less specific than frame-level or pixel-level annotations. The objective of weakly-supervised learning is to utilize these coarse labels to produce models capable of delivering fine-grained predictions, such as temporally precise action localization or detailed semantic segmentation~\cite{dong2023weakly, Long2023}. The large-scale annotation of surgical videos demands a significant investment of invaluable medical time resources. In this context, weakly supervised learning emerges as a viable solution to this bottleneck.

\section{Dataset Construction}
\label{sec:ophnet_dataset}
In this section, we detail the construction of our dataset, which involved meticulous data collection and preprocessing. YouTube was leveraged as a primary source to circumvent privacy issues while ensuring a broad representation of ophthalmic surgeries. Our selection criteria aimed to capture a wide range of video qualities and styles, specifically targeting cataract, glaucoma, and corneal surgeries due to their prevalence in clinical settings. We refined the dataset by excluding videos of inadequate quality or those depicting non-human subjects. The annotation process was designed to reflect the complex nature of eye surgeries, incorporating hierarchical classification to account for multiple conditions often treated within a single procedure. This was complemented by detailed localization annotations, delineating the distinct phases and techniques characteristic of ophthalmic operations, all undertaken by a dedicated team of ophthalmologists to ensure accuracy and relevance.

\subsection{Data Collection \& Preprocessing}
\noindent \textbf{Collection.} 
Medical data exhibit unique privacy considerations and tend to be of smaller data volumes, characteristics that are particularly salient in the case of video data. By capitalizing on the wealth of surgical videos available on YouTube, we are able to obviate potential ethical and privacy concerns while concurrently enabling the rapid procurement of a substantial corpus of videos for further screening~\cite{fair-use, YouTubeCopyrightExceptions, caba2015activitynet,kay2017kinetics}. To fulfill this objective, we deploy text-based search algorithms to probe each surgery on YouTube, obtaining videos with titles that incorporate the requisite surgical keywords. We select cataract, glaucoma, and corneal surgery—three of the most commonly performed ophthalmic surgeries in actual clinical environments—as the central subjects of our research. To expand our video collection, we bolster our search queries by integrating synonyms and abbreviations associated with each type of ophthalmic surgery. For instance, \emph{cataract surgery} encompasses various types: \emph{Phacoemulsification (abbr. PHACO)}, \emph{Intraocular Lens implantation (abbr. IOL)}, and \emph{Extracapsular Cataract Extraction (abbr. ECCE)}, etc.

\begin{figure}[t!]
\centering
\begin{overpic}[width=\textwidth]{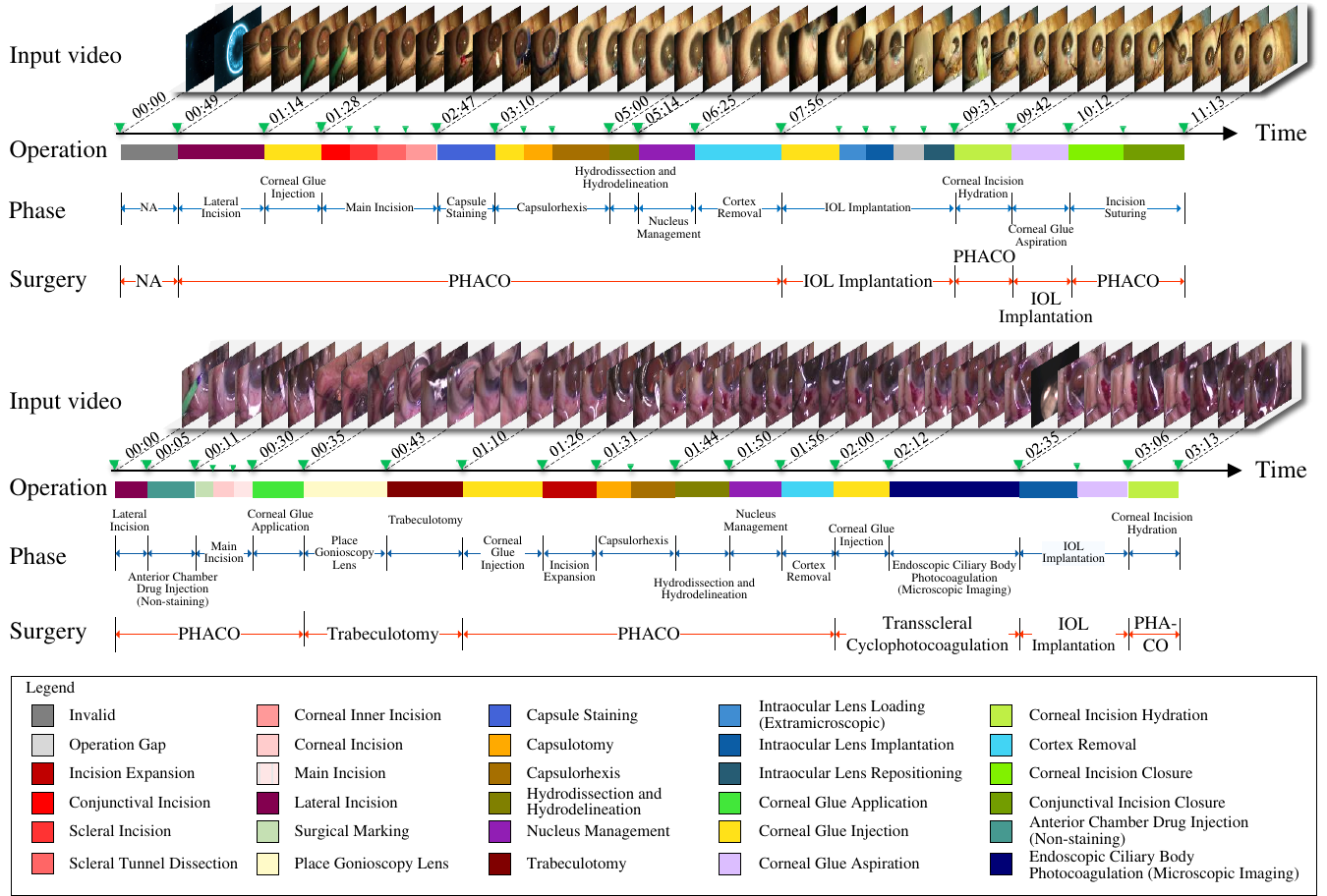}
\end{overpic}
\caption{The figure shows two combined surgical videos, \emph{PHACO + IOL implantation} and \emph{PHACO + Trabeculotomy + Transscleral Cyclophotocoagulation + IOL implantation}. For each frame marked in color, we provide time-boundary annotations at surgical, phase and operation levels.}
\label{localization}
\end{figure}

\begin{figure*}[t!]
\centering
\subfloat[]{
    \includegraphics[width=0.42\textwidth]{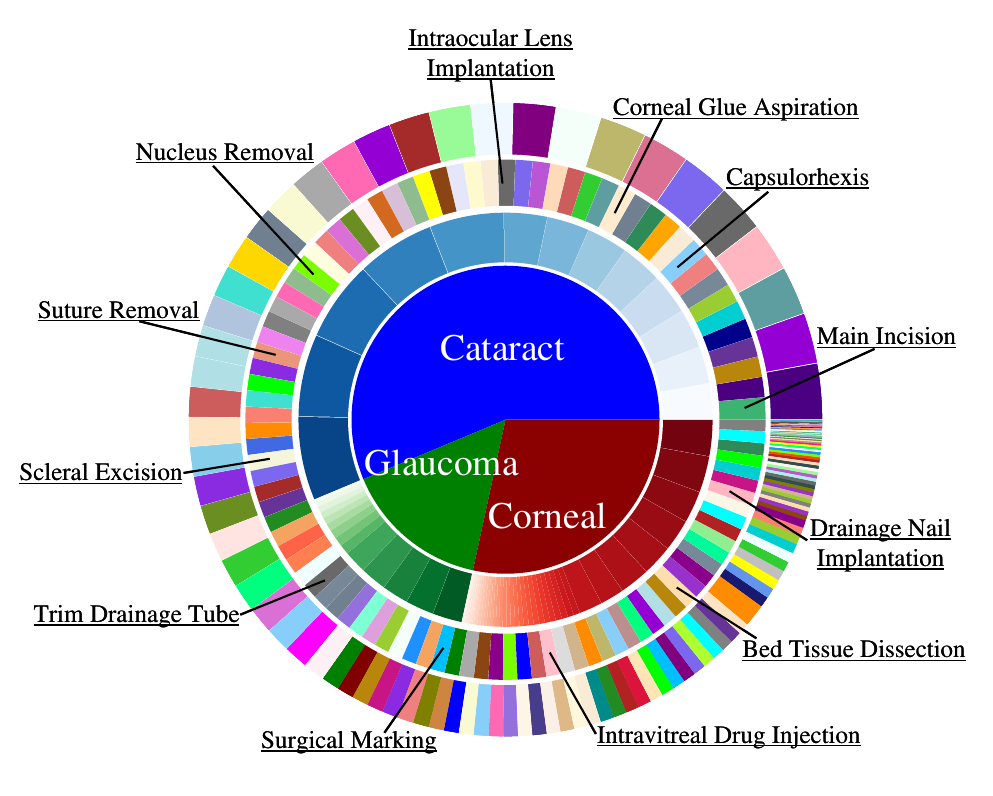}
}
\subfloat[]{
\includegraphics[width=0.54\textwidth]{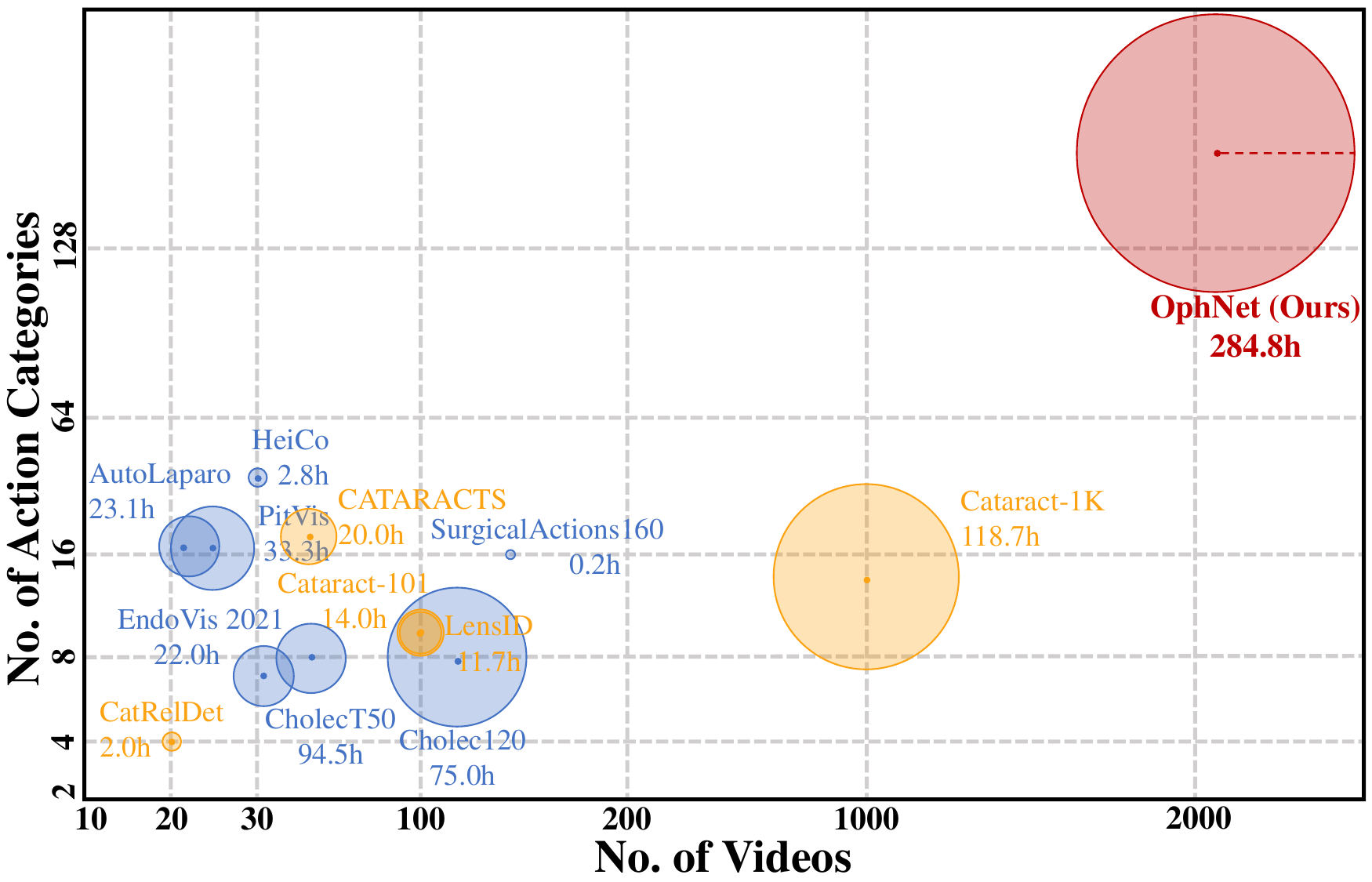}
}
\quad
\subfloat[]{
\includegraphics[width=\textwidth]{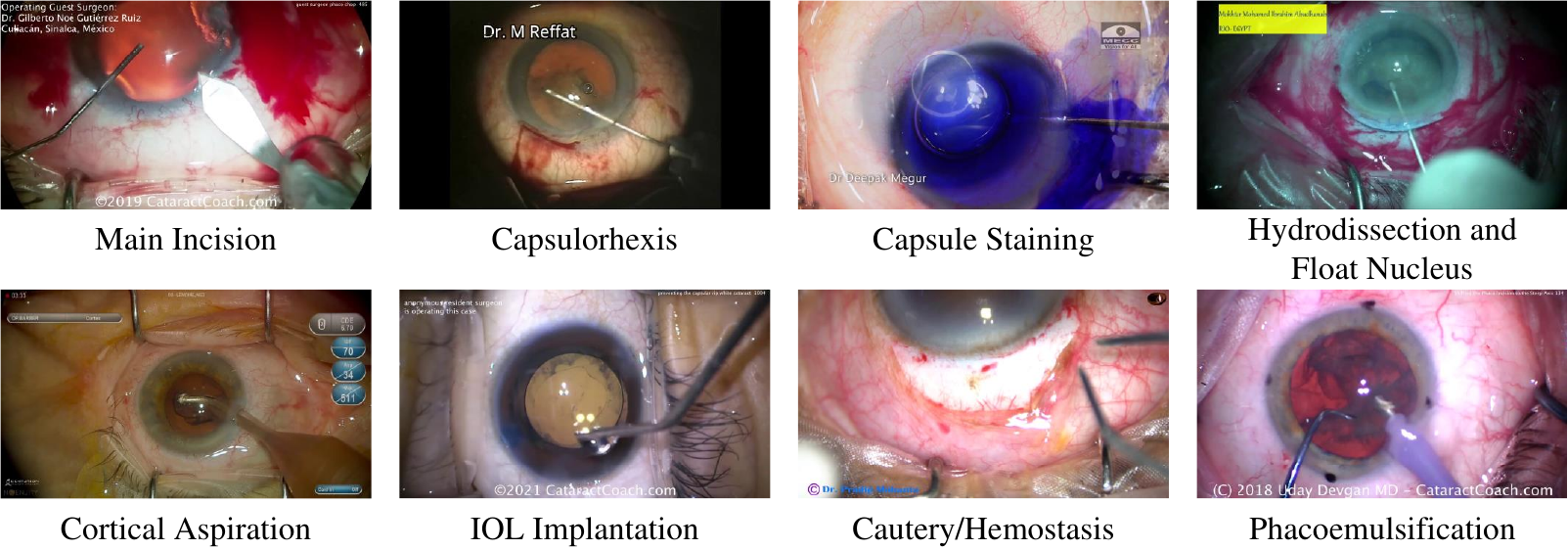}
}
\caption{OphNet's composition, comparison with other datasets for the same task, and some phase examples: (a) an overview of the composition ratios at the levels of surgery, phase, and operation; (b) comparison among existing open-source \usebox{\bluecircle}~laparoscopic \& endoscopic, and \usebox{\orangecircle}~ophthalmic microscope workflow analysis video datasets and \usebox{\redcircle}~our OphNet. OphNet stands as the largest real-world video dataset for ophthalmic surgical workflow understanding, featuring the highest number of videos, longest duration, and diverse categories of surgeries and phases; (c) eight phase examples in OphNet.}
\label{dataset_sta}
\end{figure*}

\noindent \textbf{Preprocessing.} 
Given the nature of text-based retrieval and the varying quality, style, and filming methods of surgical videos influenced by different YouTube sources, some videos may exhibit poor quality or deviations in surgical representation. We filter out low-resolution videos, black-and-white color schemes, and animated demonstrations. Furthermore, surgical videos featuring non-human eyes, such as pig eyes, rabbit eyes, or pseudo-human eyes, are also outside the scope of our annotation. The data preprocessing was jointly completed by five professionals with experience in ophthalmology for initial screening, followed by a review conducted by six attending ophthalmologists.

\subsection{Data Annotation} \label{annotation}
\noindent \textbf{Classification Annotation.} 
Different from the single-label classification in most natural video datasets~\cite{sigurdsson2016hollywood, caba2015activitynet, ZhXuCoAAAI18, gu2018ava, soomro2012ucf101}, in ophthalmic surgery, physicians typically consider and implement multiple types of surgeries and phases based on the specific conditions and needs of the patient. This multifaceted approach stems from the complexity of the eye as an organ, where numerous diseases often coexist. Consequently, multiple ocular problems may need to be addressed in a single surgery. As shown in Fig.~\ref{localization}, a patient with cataracts may also have glaucoma, necessitating the simultaneous treatment of both conditions within one surgery. Hence, we initially classified all videos based on the primary surgical categories into three principal groups: cataract, glaucoma, and corneal surgeries. Subsequently, these were further allocated for the detailed annotation of primary and secondary surgery. It is important to note that there exists only a single type of primary surgery, whereas multiple secondary surgery may be present. The categorization and annotation were meticulously completed by a team comprised of 8 experienced ophthalmologists.

\noindent \textbf{Hierarchical Localization Annotation.} 
\label{Localization-Annotation}
A single surgery is often multifaceted, involving a series of intricate phases, each requiring distinct techniques and instruments, and the transition between surgeries entails high precision and coordination. Routine cataract surgery involves several steps. It begins with the administration of anesthesia, followed by a small incision in the cornea. The surgeon then creates an opening in the lens capsule and uses ultrasonic vibrations to break up and remove the cataract. Afterward, an artificial intraocular lens (IOL) is inserted into the lens capsule, and the incision is sealed without stitches. We define the phases and operation for various surgeries based on the textbook \emph{Ophthalmic Surgery: Principles and Practice}~\cite{spaeth2011ophthalmic}. To ensure the quality of localization annotations, we assign each annotator videos of 2 different types of primary surgeries. For each action time-boundary annotation, we annotate at three different levels of granularity: surgery, phase, and operation. We also employed the complete linkage algorithm~\cite{defays1977efficient} to cluster and merge various temporal boundaries into stable boundaries that received multiple agreements. It's worth noting that an individual video may have multiple separate instances of the same or different phases, thereby leading to multiple boundary definitions. A team of 15 ophthalmologists completes the localization annotation.

\subsection{Dataset Statistics and Analysis}
OphNet includes 2,278 surgical videos (284.8 hours), demonstrating 66 different types of ophthalmic surgeries: 13 types of cataract surgery, 14 types of glaucoma surgery, and 39 types of corneal surgery. There are 102 phases and 150 operations for recognition, detection and prediction tasks, summarized in Tab.~\ref{dataset_comparison}. Over 77\% of videos have high-definition resolutions of 1280 $\times$ 720 pixels or higher. To facilitate algorithm development and evaluation, we selected 523 videos for localization annotations. Additionally, we trim videos according to annotated action boundaries, resulting in 7,320 phase instances and 9,795 operation instances, totaling 51.2 hours. The average duration of trimmed videos is 32 seconds, while untrimmed videos average 337 seconds. 

\begin{figure*}[t!]
\centering
\subfloat[phase statistics]{
    \includegraphics[width=0.48\textwidth]{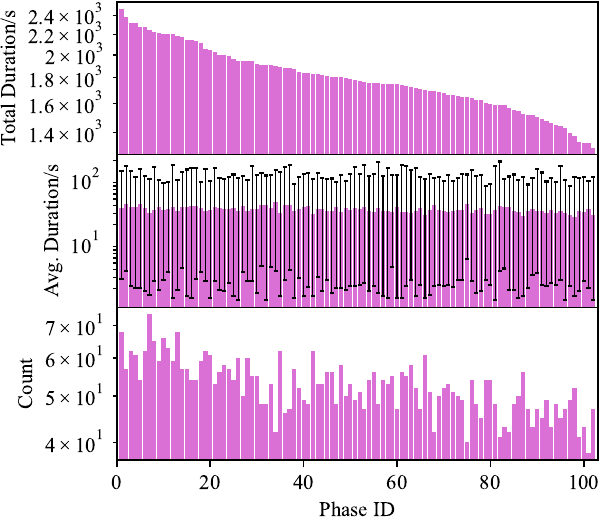}
}
\subfloat[operation statistics]{
\includegraphics[width=0.48\textwidth]{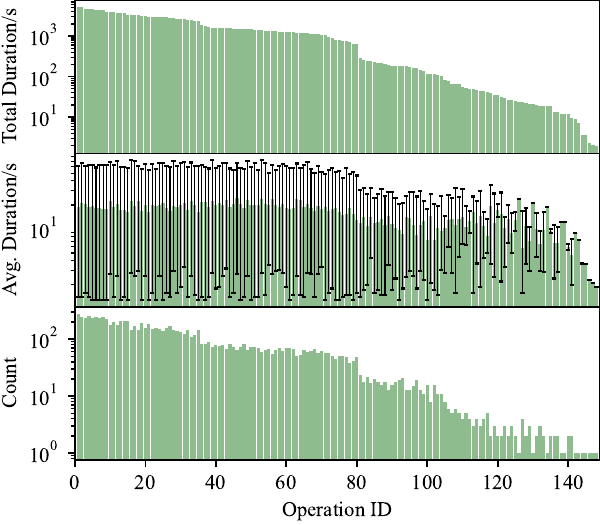}
}
\caption{We present the data statistics of trimmed videos at the levels of phase and operation, including the number of trimmed videos, average duration, and total duration. The IDs and corresponding names can be found in the appendix.}
\label{dataset_sta2}
\end{figure*}

\section{Experiments}
\label{sec:experiments}
In our study, we explore four potential tasks using the OphNet dataset: 1) primary surgery presence recognition, 2) phase and operation recognition, 3) phase localization, and 4) phase anticipation. To establish robust baselines, we employ state-of-the-art models known for their effectiveness in human action recognition, detection and anticipation. For each task, we provide a detailed problem formulation and evaluate the baseline models‘ performance. Our findings offer valuable insights into video understanding within sequences and fine granularity, contributing to the field's knowledge of video tasks in medical contexts.

\subsection{Surgery Presence, Phase and Operation Recognition}
\label{sec:experiments1}
\noindent \textbf{Task Description.}
Surgery presence recognition focuses on identifying various surgical types in untrimmed videos through weakly supervised methods. This requires the model to discern and capture distinct surgical action features within extensive video footage. In OphNet, all types of surgeries occurring in each video are annotated, but only a subset of the videos have time-boundary annotations, as detailed in Sec.~\ref{Localization-Annotation}. We identify the primary surgery based on the key objectives and durations of the surgeries. To streamline the process, our experiments are limited to recognizing the presence of these primary surgeries. Additionally, phase recognition segments the surgery into distinct phases using visual cues and movements, such as incision, lens removal, and implantation. Operation recognition involves identifying finer-grained surgical actions.

\begin{wraptable}{r}{7cm}
\tiny
\resizebox{\linewidth}{!}{%
\begin{tabular}{lcccccccccccc}
\toprule
&\multicolumn{8}{c}{\textbf{Primary Surgery Classification}} \\ \cmidrule(r){2-9}
\multirow{2}{*}{\textbf{Baselines}} & \multicolumn{2}{c}{Cataract} & \multicolumn{2}{c}{Glaucoma} & \multicolumn{2}{c}{Cornea} &
\multicolumn{2}{c}{All} \\ \cmidrule(r){2-3} \cmidrule(r){4-5} \cmidrule(r){6-7} \cmidrule(r){8-9} 
 & Top-1  & Top-5  & Top-1  & Top-5 & Top-1  & Top-5  & Top-1 & Top-5  \\
\midrule
I3D~\cite{i3d}& 38.9 &	86.3 & 43.6 & 71.7 & 42.7 &	78.2 & 29.8 & 53.2 & \\
SlowFast~\cite{feichtenhofer2019slowfast} & 45.3 & 82.1 & 44.6 & 72.3 & 45.5 & 77.3 & 27.2 & 54.4 \\
X3D~\cite{li2022mvitv2} & 42.1 & 87.4 & 44.6 & 74.2 & 43.8 & 76.6 & 28.5 & 62.7 \\
MViT V2~\cite{feichtenhofer2020x3d} & 43.2 & 84.2 &	45.5 & 81.8 & 45.5 & 75.9 &	29.1 & 60.1\\
\midrule
I3D~\cite{i3d}& 36.8  &	 84.7 & 48.2  &  81.8 &  48.6 & 76.5 & 27.2 & 50.6 \\
SlowFast* & 49.0 & 83.7 & 47.3 & 80.9 & 49.2 & 75.6 & 27.2 & 50.6 \\
X3D* & 47.4 & 86.3 & 46.4 & 81.5 & 48.3 & 78.2 & 35.4 & 61.4 \\
MViT V2* & 44.2 & 85.3 & 49.1 & 81.8 & 47.7 & 77.3 & 28.5 & 63.3\\
\midrule
$\text{X-CLIP}_{\text{16}}$~\cite{XCLIP} & 58.5 & \textbf{94.7} & 51.8 & \textbf{92.8} & \textbf{61.4} & \textbf{88.6} & 40.5 & 79.0 \\

$\text{X-CLIP}_{\text{32}}$ & \textbf{60.6} & 92.6 &  \textbf{53.5} & 83.7 & 56.8 & 84.1 & \textbf{58.9} & \textbf{81.0}
\\

$\text{ViFi-CLIP}_{\text{16}}$~\cite{hanoonavificlip} & 59.6 & 88.3 & 52.4 & 80.2 & 61.4 & 81.8 & 58.9 & 79.8 & 
\\

$\text{ViFi-CLIP}_{\text{32}}$ & 59.6 & 88.3 & 51.5 & 84.8 & 50.0 & 75.0 & 56.3 & 77.2 
\\
\bottomrule
\end{tabular}%
}
\caption{Per-class Top-1 and Top-5 accuracy (\%) for the primary surgery presence recognition on untrimmed videos. The best performance for each split has been highlighted in \textbf{bold}.}
\label{tab:surgery_classification}
\end{wraptable}

\noindent \textbf{Baselins.} We compare the performance of I3D~\cite{i3d}, SlowFast~\cite{feichtenhofer2019slowfast}, X3D~\cite{feichtenhofer2020x3d}, and MViT V2~\cite{li2022mvitv2} models on this task. These models are evaluated in two versions: 1) random initialization training and 2) pre-training with weights from Kinetics 400~\cite{kay2017kinetics}, which is a human action recognition dataset. In addition, we also explored the classification performance of X-CLIP~\cite{XCLIP} and ViFi-CLIP~\cite{hanoonavificlip}, two CLIP~\cite{radford2021learning}-based models. We also compared the effects of different numbers of input frames on the models' performance, where the subscript \emph{\_16} represents an input of 16 frames, and \emph{\_32} represents an input of 32 frames.

\noindent \textbf{Setup.}
The dataset was randomly partitioned to ensure a balanced representation of examples for each surgery category. Specifically, we allocated 70\% of the data for training (1,449 surgical videos), 10\% for validation (205 surgical videos), and 20\% for testing (424 surgical videos). For the phase and operation recognition experiments, we followed the same settings, with 70\% of the data used for training (5,024 phase segments, 6,856 operations segments), 10\% for validation (730 phase segments, 975 operations segments), and 20\% for testing (1,566 phase segments, 1,964 operations segments). The input for the surgery presence recognition experiment is untrimmed videos, while for the phase and operation recognition experiments, the input consists of trimmed segments. In all classification experiments, we set up the analysis and comparison from four perspectives: cataract surgery, glaucoma surgery, corneal surgery, and all surgical videos.

\begin{figure}[t!]
\centering
\begin{overpic}[width=\textwidth]{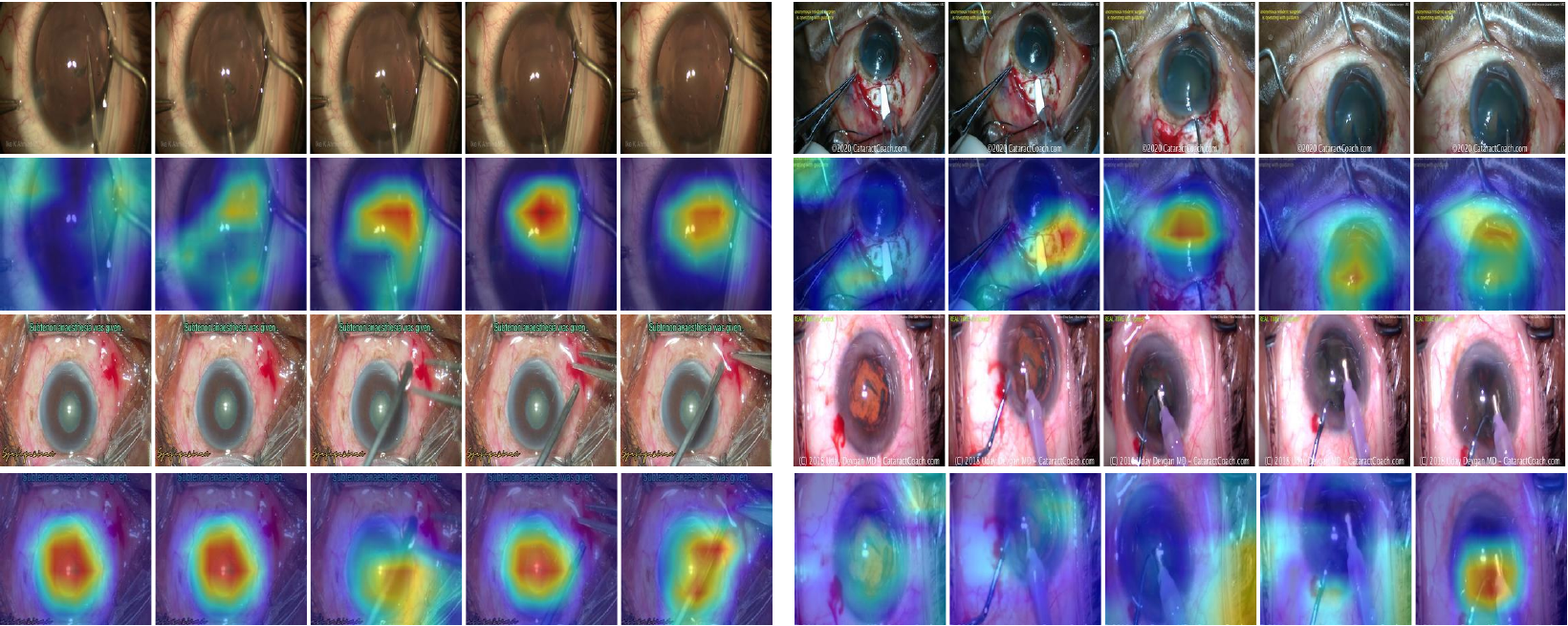}
\end{overpic}
\caption{Attention map visualizations of ViFi-CLIP~\cite{hanoonavificlip} on four examples from OphNet's test set in the phase recognition task.}
\label{heatmap}
\end{figure}

\noindent \textbf{Results.} 
We summarize the results in Tab.~\ref{tab:surgery_classification} and Tab.~\ref{tab:phase_operation_classification}. In primary surgery classification, X-CLIP~\cite{XCLIP} achieved the highest overall Top-1 accuracy at 58.9\%, leading in cataract and glaucoma surgeries with accuracies of 60.6\% and 53.5\% respectively. For corneal surgeries, X-CLIP also recorded the highest Top-1 and Top-5 accuracies of 61.4\% and 88.6\%. In phase classification, ViFi-CLIP~\cite{hanoonavificlip} led with the highest Top-1 accuracy, particularly in cataract surgeries at 75.9\%, and exhibited the best performance in both glaucoma and corneal surgeries. Furthermore, in operation classification, ViFi-CLIP outperformed all other models in all categories, especially noted in cataract and corneal surgeries with Top-1 accuracies of 75.1\% and 83.7\% and Top-5 accuracies of 93.8\% and 85.2\% respectively. Overall, ViFi-CLIP showed superior performance in both phase and operation classifications across various surgery types. Besides, in the phase and operation classification experiments, a higher number of input frames generally had a positive effect on the model. In Fig.~\ref{heatmap}, we present the heatmap visualization of four examples from the test set of phase recognition experiments using the ViFi-CLIP model. It can be observed that in a series of frame images, the model focuses on surgical instruments and the operated eye area, which is consistent with human experience. The instruments used in the same phase of ophthalmic surgery are often similar.

\subsection{Phase Localization}
\noindent \textbf{Task Description.}
Phase localization in ophthalmic surgical workflow analysis refers to the task of pinpointing the exact moments or time intervals within a surgical video where specific phases of the surgery begin and end. This involves the detailed temporal segmentation of the entire surgical procedure into its constituent phases based on visual cues, surgeon's actions, and the progression of the surgery. The objective of phase localization is to accurately identify the start and end times of different surgical stages, such as pre-operative preparation, incision, and removal of the lens facilitating a granular and precise understanding of the surgery timeline. This task is crucial for detailed surgical documentation, efficient surgical training, and the development of targeted interventions during specific stages of the surgery, enhancing overall surgical management and post-operative analysis.

\begin{table*}[t!]
\tiny
\begin{center}
\tabcolsep=0.01cm
\setlength{\abovecaptionskip}{1pt}
\resizebox{\linewidth}{!}{
\begin{tabular}{lcccccccccccccccccccc}
\toprule
&\multicolumn{8}{c}{\textbf{Phase Classification}}  
&\multicolumn{8}{c}{\textbf{Operation Classification}}  \\ \cmidrule(r){2-9} \cmidrule(r){10-17}
\multirow{2}{*}{\textbf{Baselines}} & \multicolumn{2}{c}{Cataract} & \multicolumn{2}{c}{Glaucoma} & \multicolumn{2}{c}{Cornea} &
\multicolumn{2}{c}{All} &
\multicolumn{2}{c}{Cataract} & \multicolumn{2}{c}{Glaucoma} & \multicolumn{2}{c}{Cornea} &
\multicolumn{2}{c}{All} \\ \cmidrule(r){2-3} \cmidrule(r){4-5} \cmidrule(r){6-7} \cmidrule(r){8-9} \cmidrule(r){10-11} \cmidrule(r){12-13} \cmidrule(r){14-15} \cmidrule(r){16-17}
 & Top-1  & Top-5  & Top-1  & Top-5 & Top-1  & Top-5  & Top-1 & Top-5 & Top-1  & Top-5  & Top-1  & Top-5 & Top-1  & Top-5  & Top-1  & Top-5 \\
\midrule
I3D~\cite{i3d}& 
27.2 & 55.7 & 24.1 & 57.5 & 18.9 & 52.1 & 25.7 & 58.2 & 26.8 & 54.9 & 23.5 & 56.0 & 18.0 & 51.2 & 25.0 & 57.1 \\
SlowFast~\cite{feichtenhofer2019slowfast} 
& 26.5 & 56.5 & 23.1 & 56.5 & 24.2 & 49.1 & 26.7 & 60.1 & 25.8 & 55.2 & 22.9 & 45.9 & 23.5 & 48.5 & 26.0 & 59.0 \\
X3D~\cite{li2022mvitv2} 
& 27.0 & 58.3 & 21.0 & 55.5 & 21.8 & 28.5 & 26.6 & 62.3 & 26.4 & 47.2 & 20.5 & 44.6 & 21.3 & 27.8 & 26.1 & 61.5 \\
MViT V2~\cite{feichtenhofer2020x3d} 
& 26.2 & 54.9 & 21.0 & 53.4 & 26.0 & 46.8 & 27.0 & 59.8 & 25.9 & 43.8 & 20.5 & 42.7 & 25.5 & 45.9 & 26.5 & 58.9 \\
\midrule
I3D* 
& 29.5 & 68.9 & 22.1 & 58.6 & 26.0 & 50.9 & 30.2 & 71.2 & 28.8 & 67.5 & 21.8 & 47.9 & 25.7 & 50.3 & 29.5 & 60.0 \\
SlowFast* 
& 30.6 & 72.3 & 25.2 & 54.7 & 30.7 & 59.8 & 31.7 & 61.8 & 29.9 & 71.1 & 24.8 & 43.9 & 29.5 & 58.7 & 30.5 & 60.9 \\
X3D* 
& 27.2 & 72.9 & 22.1 & 59.6 & 30.7 & 61.5 & 33.5 & 63.2 & 26.5 & 71.8 & 21.7 & 48.8 & 29.9 & 60.2 & 32.8 & 62.1 \\
MViT V2* 
& 34.2 & 76.5 & 23.3 & 52.0 & 38.4 & 65.1 & 28.3 & 60.2 & 33.5 & 75.2 & 22.8 & 41.5 & 37.9 & 64.0 & 27.8 & 59.5 \\
\midrule
$\text{X-CLIP}_{\text{16}}$~\cite{XCLIP}  & 68.3 & 92.2 & 47.3 & 89.8 & 53.0 & 77.4 & 63.4 & 85.3 & 67.5 & 91.0 & 46.5 & 78.9 & 82.2 & 76.1 & 62.5 & 84.0 \\
$\text{X-CLIP}_{\text{32}}$ & 69.1 & \textbf{94.0} & 48.7 & 81.7 & 54.8 & 80.4 & 62.7 & 85.8 & 68.0 & 93.0 & \textbf{47.9} & \textbf{80.5} & 84.0 & 79.5 & 62.0 & 84.7 \\
$\text{ViFi-CLIP}_{\text{16}}$~\cite{hanoonavificlip} & \textbf{75.9} & 93.7 & 40.4 & 85.4 & 66.6 & 81.6 & 66.1 & \textbf{88.4} & 74.5 & 92.5 & 42.8 & 74.5 & \textbf{85.0} & 80.5 & \textbf{65.0} & \textbf{87.5} \\

$\text{ViFi-CLIP}_{\text{32}}$ 
& 73.0 & 92.9 & \textbf{49.6} & \textbf{82.7} & \textbf{57.7} & \textbf{81.6} & \textbf{68.4} & 87.2 & \textbf{75.1} & \textbf{93.8} & 43.2 & 80.2 & 83.7 & 85.2 & 64.8 & 86.5 \\
\bottomrule
\end{tabular}}
\end{center}
\caption{Per-class Top-1 and Top-5 accuracy (\%) for the primary surgery presence recognition on untrimmed videos and phase recognition on trimmed videos. * denotes the initialization from the model pre-trained on Kinetics 400~\cite{kay2017kinetics}. For the two CLIP models, we chose ViT-B/16 as the backbone and compared the performance of two different input frame numbers, 16 and 32. The best performance for each split has been highlighted in \textbf{bold}.} 
\label{tab:phase_operation_classification}
\end{table*}

\begin{table}[t!]
\centering
\begin{minipage}[t]{0.59\textwidth}
\centering
\tiny
\tabcolsep=0.01cm
\setlength{\abovecaptionskip}{6pt}
\resizebox{\textwidth}{!}{%
\begin{tabular}{llccccc}
    \toprule
    & \multicolumn{5}{r}{\textbf{mAP (\%)}} \\
    \cmidrule(r){3-7}
    \multirow{-3}{*}{\textbf{Baselines}} & \multirow{-3}{*}{\textbf{Backbones}} & 0.1 & 0.3 & 0.5 & 0.7 & Avg. \\
    \midrule
     & CSN~\cite{tran2019video} & 53.7 & 50.1 & 40.6 & 24.5 & 42.5 \\
   ActionFormer~\cite{zhang2022actionformer} & SwinViviT~\cite{liu2021video} & 59.3 & 54.7 & 43.3 & 26.3 & 46.4 \\
    & SlowFast~\cite{feichtenhofer2019slowfast} & 60.0 & 55.9 & 45.1 & 26.0 & 47.5 \\
    \midrule
     & CSN & 56.1 & 53.0 & 43.1 & 29.4 & 46.2 \\
   TriDet~\cite{shi2023tridet} & SwinViviT & 61.0 & \textbf{57.1} & \textbf{47.1} & \textbf{33.1} & \textbf{50.4} \\
    & SlowFast & \textbf{61.3} & 56.0 & 45.6 & 30.4 & 48.6 \\
    \bottomrule
\end{tabular}%
}
\caption{The results for phase detection. ActionFormer and TriDet are state-of-the-art models for human action detection tasks, and we use three different backbones for feature extraction and report mAP at the IoU thresholds of [0.1:0.2:0.9]. Average mAP is computed by averaging different IoU thresholds. The best performance for each split has been highlighted in \textbf{bold}.}
\label{tab:phase_detection}
\end{minipage}%
\hfill 
\begin{minipage}[t]{0.39\textwidth}
\centering
\tiny
\tabcolsep=0.01cm
\setlength{\abovecaptionskip}{6pt}
\resizebox{\textwidth}{!}{%
\begin{tabular}{lccccc}
    \toprule
    & \multicolumn{5}{c}{\textbf{Top-1 Acc. (\%)}} \\
    \cmidrule(r){2-6}
    \multirow{-3}{*}{\textbf{Baselines}}  & 0.1 & 0.3 & 0.5 & 0.7 & Avg. \\
    \midrule
    I3D~\cite{i3d} &  26.5 & 42.2 & 49.8 & 51.3 & 47.3 \\
    SlowFast~\cite{feichtenhofer2019slowfast}  &  25.4 & 42.6 & 48.9 & 52.2 & 47.2 \\
    MViT V2~\cite{feichtenhofer2020x3d} & 25.6 & 43.7 & 49.3 & 52.3 & 47.5 \\
    \midrule
    I3D* &  27.3 & 43.5 & 50.1 & 51.4 & 47.6 \\
    SlowFast* & 27.5 & 43.2 & 49.9 & \textbf{52.3} & 47.8  \\
    MViT V2* & \textbf{27.8} & \textbf{43.8} & \textbf{50.5} & 51.7 & \textbf{48.2}  \\
    \bottomrule
\end{tabular}%
}
\caption{The results for phase anticipation. We report top-1 accuracy at the observation ratios [0.1:0.2:0.9]. Average top-1 accuracy is computed by averaging different observation ratios. The best performance for each split has been highlighted in \textbf{bold}.}
\label{tab:phase_prediction}
\end{minipage}
\end{table}

\noindent \textbf{Baselines.}
We conducted the experiments using phase-level labels and used two baseline models, ActionFormer~\cite{zhang2022actionformer} and TriDet~\cite{shi2023tridet}, with backbone networks configured as CSN~\cite{tran2019video}, SwinViviT~\cite{liu2021video}, and SlowFast~\cite{feichtenhofer2019slowfast}. Data split follows the setup of the primary surgery classification experiment.

\noindent \textbf{Setup.}
We excluded \emph{Operation Gap} and \emph{Invalid}, and filtered out tags with fewer than 20 segments. We used two baseline models, ActionFormer~\cite{zhang2022actionformer} and TriDet~\cite{shi2023tridet}, with backbone networks configured as CSN~\cite{tran2019video}, SwinViviT~\cite{liu2021video}, and SlowFast~\cite{feichtenhofer2019slowfast}. To extract features from the videos, we first extracted RGB frames from each video at a rate of 25 frames per second. We also extracted optical flow using the TV-L1~\cite{horn1981determining,lucas1981iterative} algorithm. We then fine-tuned an I3D~\cite{i3d} model that had been pre-trained on the ImageNet~\cite{deng2009imagenet} dataset, and used it to generate features for each RGB and optical flow frame. Because each video has a variable duration, we performed uniform interpolation to generate 100 fixed-length features for each video. Finally, we concatenated the RGB and optical flow features into a 2048-dimensional embedding, which served as the input for our model.

\noindent \textbf{Results.}
The experimental results for phase localization are presented in the Tab.~\ref{tab:phase_detection}, showcasing the performance of different baseline models with various backbones in terms of mean Average Precision (mAP) at different Intersection over Union (IoU) thresholds [0.1:0.2:0.9]. The models evaluated include ActionFormer with SwinViviT and SlowFast backbones, and TriDet with CSN, SwinViviT, and SlowFast backbones. The results indicate that the TriDet model with a SwinViviT backbone outperforms other combinations, achieving the highest mAP scores across most IoU thresholds, with notable scores of 61.0\% (IoU=0.1), 57.1\% (IoU=0.3), 47.1\% (IoU=0.5), and 33.1\% (IoU=0.7), resulting in an average mAP of 50.4\%. This indicates that the TriDet model, especially when combined with the SwinViviT backbone, is particularly effective for phase localization in surgical videos. On the other hand, the TriDet model with a SlowFast backbone shows competitive performance, particularly achieving the highest mAP of 61.3\% at the lowest IoU threshold (0.1). However, it falls slightly behind in performance at higher IoU thresholds compared to the SwinViviT backbone.

\subsection{Phase Anticipation}
\noindent \textbf{Task Description.}
This task requires the analysis of real-time or recorded video data to foresee the sequence of events based on current and past surgical activities. By understanding the typical progression of ophthalmic surgeries and recognizing patterns in the surgeon's actions and the use of instruments, the system aims to forecast the next phase of the surgery, allowing for proactive preparation and response. The objective of phase anticipation is to enhance the efficiency and safety of surgical procedures by providing the surgical team with advanced notice of upcoming steps, enabling better resource allocation, timing for critical tasks, and overall coordination within the operating room. 

\noindent \textbf{Setup.}
Following previous approaches of the primary surgery classification experiment in Sec.~\ref{sec:experiments1}, We randomly mask phase sequences in the test video with different observation ratios.

\noindent \textbf{Baselines.}
We evaluate our datasets with the baseline models such as I3D~\cite{i3d}, SlowFast~\cite{feichtenhofer2019slowfast}, and MViT V2~\cite{li2022mvitv2}. For each model, we also adopted two training approaches: random initialization training and using pre-trained weights from Kinetics 400~\cite{kay2017kinetics}.

\noindent \textbf{Results.} 
The phase detection results are illustrated in Tab.~\ref{tab:phase_prediction}. The results demonstrate that the baseline models pretrained on Kinetics 400~\cite{kay2017kinetics} generally outperform their original counterparts in terms of Top-1 accuracy across different observation ratios for phase anticipation. Specifically, the modified MViT V2* model exhibits the highest improvement, achieving the best average Top-1 accuracy of 48.2\%. Moreover, while all models show increased accuracy with higher observation ratios, indicating that more observed data contributes to better performance, the consistent improvement across all ratios for the enhanced models suggests effective modifications.

\begin{figure}[t!]
\centering
\begin{overpic}[width=\textwidth]{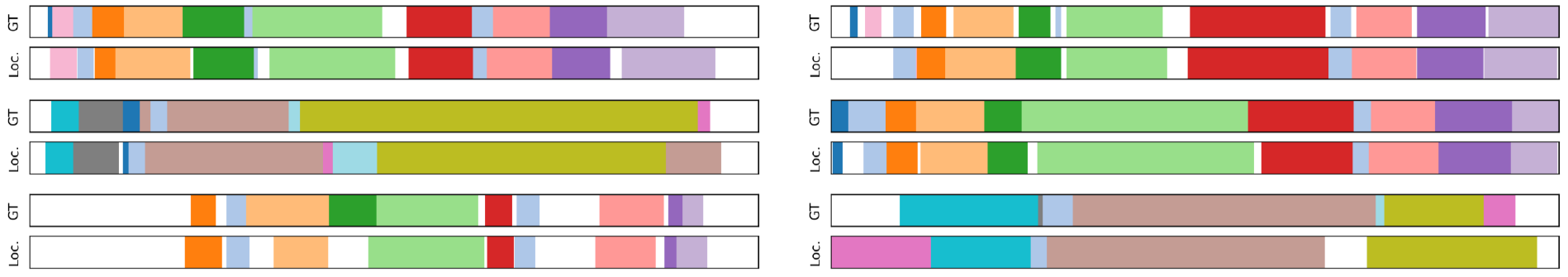}
\end{overpic}
\caption{Phase localization visualization of TriDet~\cite{shi2023tridet}. \emph{GT} represents the ground truth visualization for phases, while \emph{Loc.} visualizes the model's highest confidence phase category and the time-boundary results. Blank segments denote invalid segments or operation gaps.}
\label{loca_vis}
\end{figure}

\section{Limitations}
\label{sec:limitations}
\noindent \textbf{Dataset Bias.}
OphNet's videos are sourced from YouTube and exhibit diverse styles, clarity, and screen elements. This diversity can aid detection models in generalization but may affect their effectiveness and performance. Some videos in the dataset include subtitles or additional video windows, such as a little subtitle or watermark shown in Fig.~\ref{dataset_sta}. Similarly, additional video windows offer another perspective but can make the scene chaotic, making it harder to recognize primary surgical actions. The presence of these factors in OphNet reflects the complexity of real-world surgical environments, because an ophthalmic microscope may inherently display different windows or show parameters during recording. While they pose challenges, they also present opportunities for developing models that can better handle variability and unpredictability, which are crucial aspects of real-world surgical scenarios.

\noindent \textbf{Annotation Bias.}
OphNet is entirely annotated by ophthalmologists, there is a distinct possibility of annotation bias reflecting specific regional practices, terminologies, and interpretations. Despite the universal nature of many ophthalmic procedures, subtle differences in surgical techniques, procedural preferences, and clinical terminologies could lead to inconsistencies in how surgeries are categorized and described across different regions. For instance, the terminology used to describe certain procedures might differ, with one region referring to a procedure as \emph{anterior vitrectomy} while another uses \emph{pars plana vitrectomy}. To reduce the possibility of biases in precise annotations, we have taken great care to establish a unified definition prior to describing the surgery, phase and operation. However, potential biases arising from regional variations and individual surgical practices are inevitable.

\section{Conclusion}
\label{sec:conclusion}
In response to the current challenges in ophthalmology, a surgical field apt for automation and remote control, we introduce OphNet, a large-scale, diverse, and expert-level video benchmark for understanding ophthalmic surgical workflows. OphNet is the most extensive dataset of its kind, containing a broad range of cataract, glaucoma, and corneal surgeries and detailed annotations for distinct surgical phases. OphNet comprises 2,278 surgical videos (284.8 hours), 7,320 phase segments and 9,795 operation segments (51.2 hours), showcasing 66 different types of ophthalmic surgeries: 13 cataract, 14 glaucoma, and 39 corneal. It is annotated with 102 phases and 150 operations. With OphNet, we explored primary surgery presence recognition, phase localization and phase anticipation on untrimmed videos, phase and operation recognition on trimmed videos. We employed state-of-the-art models to establish robust baselines and provided valuable insights into video understanding within sequences and fine granularity. Our work contributes to the broader understanding of surgical video tasks in medical contexts and promotes the integration of deep learning technologies into ophthalmic surgical procedures.

\bibliographystyle{splncs04}
\bibliography{main}

\clearpage

\section{Training Details}
We conducted all experiments using 4 NVIDIA RTX3090Ti GPUs. 

\subsection{Training Details for Classification Tasks}
We employed the officially released codes to train all recognition models. The SlowFast~\cite{feichtenhofer2019slowfast}, I3D~\cite{i3d} and X3D~\cite{feichtenhofer2020x3d}models were trained for 150 epochs with a batch size of 16, using a base learning rate of 0.001. We employed a cosine decay learning rate scheduler with 34 warmup epochs. We sampled 16 frames per clip with a sampling rate of 16. For the configuration of training MViT v2~\cite{li2022mvitv2} model, we apply the base learning 0.0001, cosine decay learning rate scheduler, 200 training epochs, 30 warmup epochs, and the batch size 8. We sample 16 frames per clip with the sampling rate of 16.

\subsection{Training Details for Localization Task}
\noindent \textbf{Feature extraction.} We firstly extract the frames from each video with 25 FPS and also extract the optical flow with TV-L1~\cite{horn1981determining,lucas1981iterative} algorithm. After that, we finetune an I3D~\cite{i3d} model on Kinetics 400~\cite{kay2017kinetics}, and then use it to generate the features for each RGB and optical flow frame. Since each video has variable duration, we perform the uniform interpolation to generate 100 fixed-length features for each video. Finally, we concatenate the RGB and optical flow features into a 2048-dimensional embedding as the model input.

\noindent \textbf{Model training.} We train all the detection models with their officially released code and the default configurations. For training ActionFormer~\cite{actionformer} model, we apply the base learning rate 0.001, cosine decay learning rate scheduler, 30 training epochs, 5 warmup epochs, and the batch size 16. For training TriDet~\cite{shi2023tridet} model, we apply the base learning rate of 0.0004, step decay learning rate scheduler, 20 training epochs, and the batch size 200. For these two baseline models, we employed three different backbone network settings for performance comparison: CSN~\cite{tran2019video}, SwinViviT~\cite{liu2021video}, and SlowFast~\cite{feichtenhofer2019slowfast}.

\subsection{Training Details for Anticipation Task}
We follow the same settings as used in classification experiment.

\begin{figure*}[t!]
\centering
\includegraphics[width=0.95\linewidth]
                  {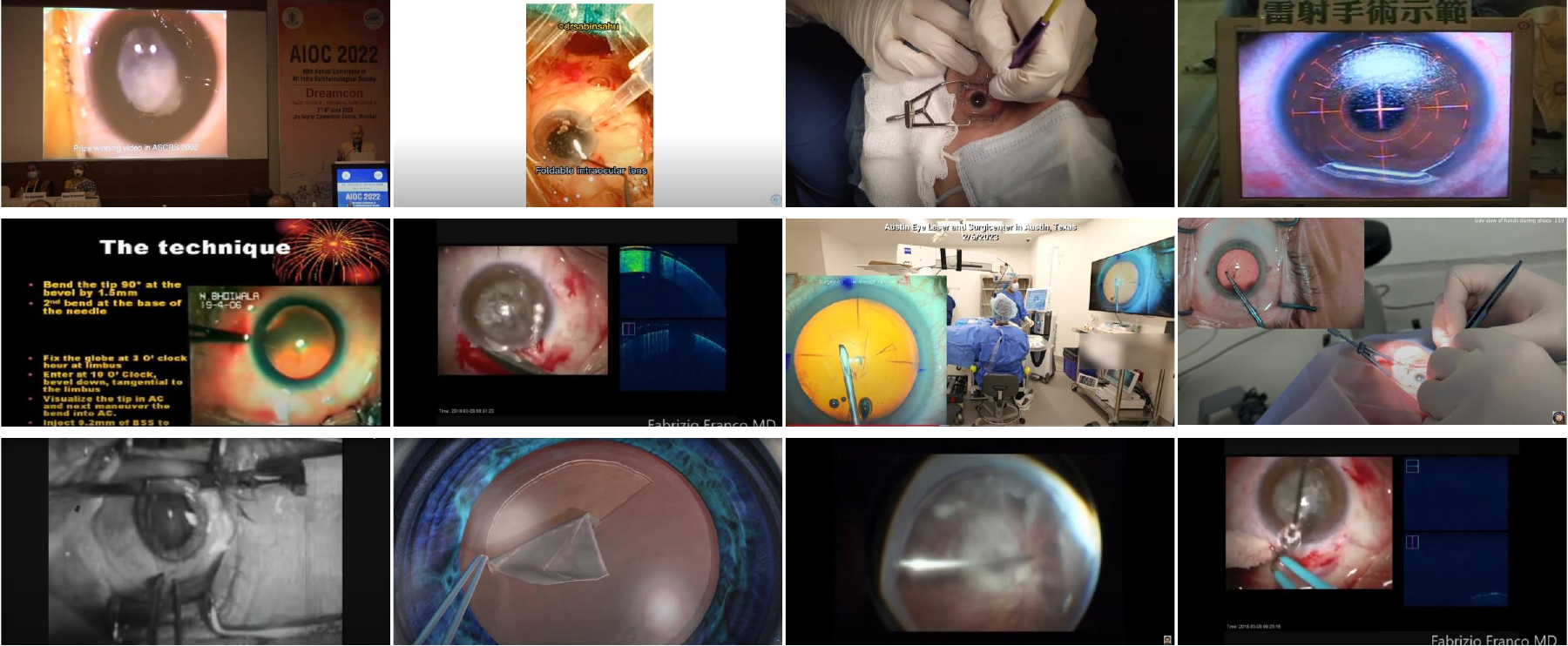}
\caption{Examples of filtered videos.}
\label{bias1}
\end{figure*}

\begin{figure*}[t!]
\centering
\includegraphics[width=0.95\linewidth]
                  {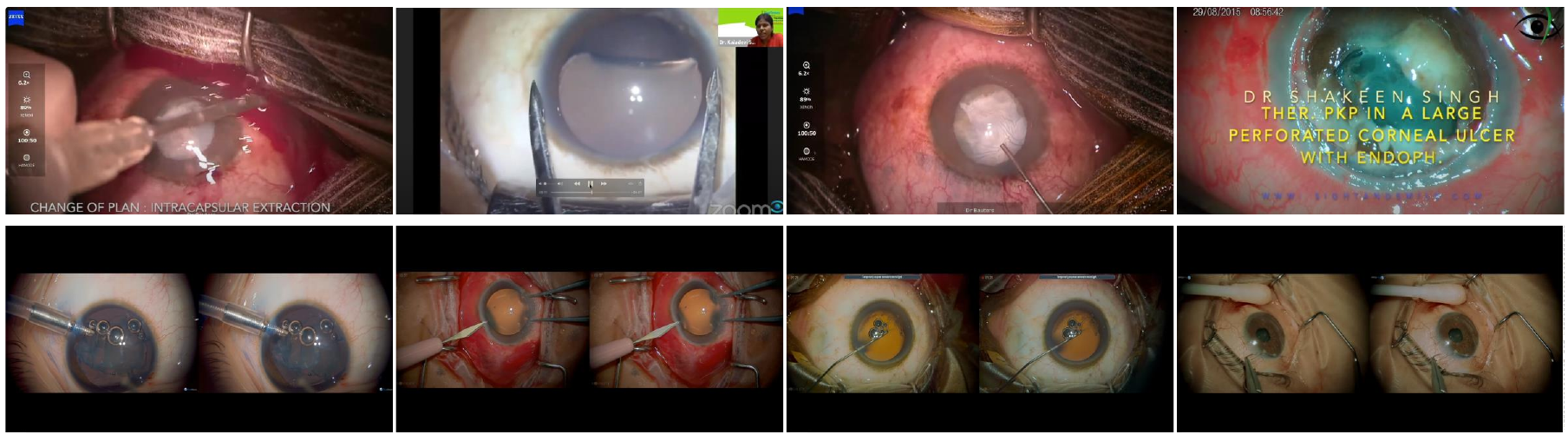}
\caption{Examples with minor flaws that were still retained.}
\label{bias2}
\end{figure*}

\section{Annotation Interface Demonstration} 
\subsection{Video Filtering}
The videos in the OphNet dataset, sourced from YouTube, exhibit a variety of styles, resolutions, and on-screen elements. To ensure quality and relevance, we filtered out videos that do not provide a microscopic perspective (first row of Fig.~\ref{bias1}), as well as those with subtitles, additional video windows, or watermarks occupying a significant portion of the frame (second row of Fig.~\ref{bias1}). Furthermore, videos depicting unrealistic animations, suffering from poor resolution, displaying grayscale images, or containing OCT imagery (third row of Fig.~\ref{bias1}) were also excluded. However, we retained videos with minimal on-screen text or watermarks (first row of Fig.~\ref{bias2}). Additionally, 3D videos recorded using binocular microscopes were preserved, albeit processed to retain only the left-eye perspective in our dataset.

\subsection{Classification Annotation Interface}
In this stage, we categorize the videos into valid and invalid videos through keypresses, with valid videos further classified based on their primary surgical type. Initially, an attending ophthalmologist categorizes the videos into three types: cataract surgery, glaucoma surgery, and corneal surgery. These are then further distributed for filtering and classification annotation, with each individual responsible for one of the three major surgeries.

\begin{figure*}[t!]
\centering
\subfloat[Video filtering and surgery classification annotation interface.]{
    \includegraphics[width=0.95\textwidth]{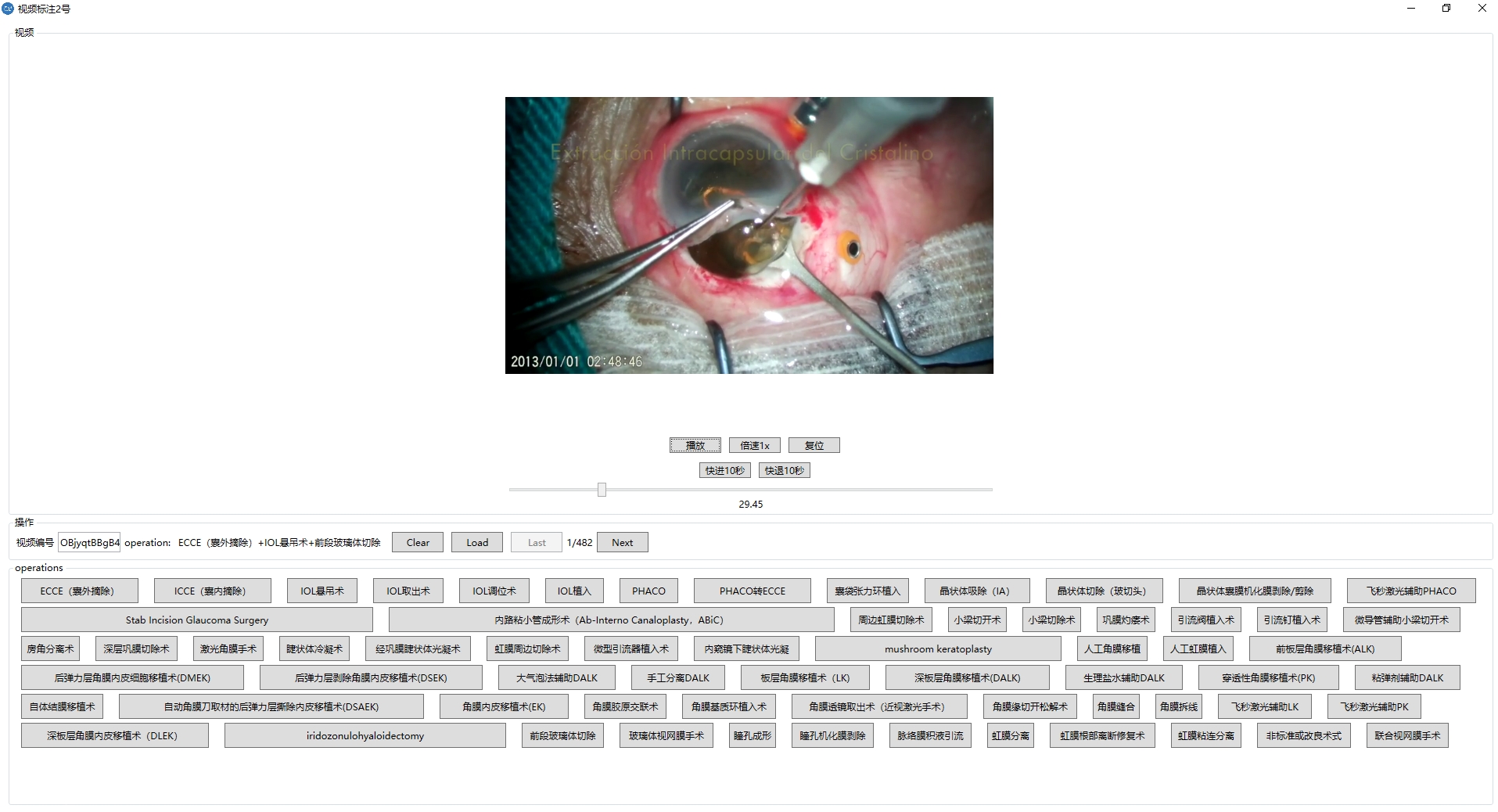}
}
\quad
\subfloat[Hierarchical temporal localization annotation interface for surgery, phase, and operation]{
\includegraphics[width=0.95\textwidth]{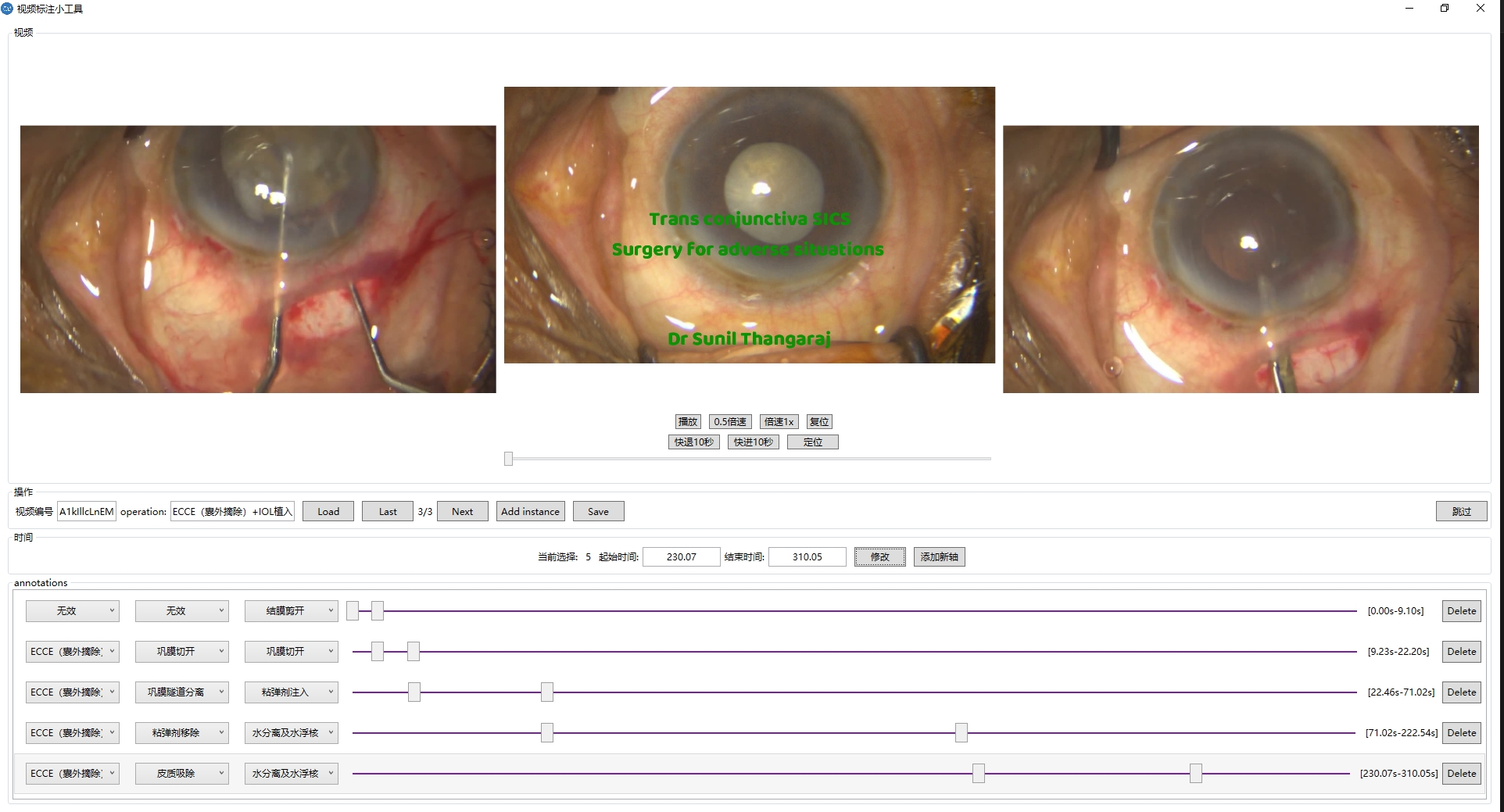}
}
\caption{Annotation Interface Design}
\label{interface}
\vspace{-0.75 cm}
\end{figure*}

\subsection{Hierarchical Localization Annotation Interface}
We have designed an interface that supports three levels of annotation: surgery, phase, and operation, and is easy to operate and modify later. The main window plays the video (with features such as speed adjustment, fast forward, rewind, and pause), while the left and right sub-windows display the corresponding frames for the start and end times of the current annotated segment. Additionally, it supports functions such as automatic time positioning and instance insertion.



\section{Dataset Bias} 
\noindent \textbf{Dataset Bias.}
OphNet's videos are sourced from YouTube and exhibit diverse styles, clarity, and screen elements. This diversity can aid detection models in generalization but may affect their effectiveness and performance. Some videos in the dataset include subtitles or additional video windows, such as watermark shown in Fig.~\ref{bias1}. Similarly, additional video windows offer another perspective but can make the scene chaotic, making it harder to recognize primary surgical actions. The presence of these factors in OphNet reflects the complexity of real-world surgical environments, because an ophthalmic microscope may inherently display different windows or show parameters during recording. While they pose challenges, they also present opportunities for developing models that can better handle variability and unpredictability, which are crucial aspects of real-world surgical scenarios.

\noindent \textbf{Annotation Bias.}
OphNet is entirely annotated by ophthalmologists, and while this ensures a high level of expertise, it also introduces the possibility of annotation bias reflecting specific regional practices, terminologies, and interpretations. Despite the universal nature of many ophthalmic procedures, subtle differences in surgical techniques, procedural preferences, and clinical terminologies could lead to inconsistencies in how surgeries are categorized and described across different regions. For instance, the technique for cataract extraction may vary between phacoemulsification in one region and manual small incision cataract surgery in another, leading to differences in the annotation of surgical phases and operations. Similarly, the terminology used to describe certain procedures might differ, with one region referring to a procedure as "anterior vitrectomy" while another uses "pars plana vitrectomy." To reduce the possibility of biases in precise annotations, we have taken great care to establish a unified definition prior to describing the surgery, phase, and operation. However, potential biases arising from regional variations and individual surgical practices are inevitable. Recognizing these potential biases is crucial for the users of OphNet, as it allows for a more nuanced interpretation of the data and its applicability to different clinical settings. Future work could involve expanding the annotation team to include ophthalmologists from diverse geographical regions and surgical backgrounds, further mitigating the impact of regional and individual biases on the dataset.

\section{OphNet's Extension}
\noindent \textbf{Multi-Surgery Recognition.}
In the realm of surgical procedures, obtaining large-scale, finely annotated video datasets is a formidable challenge due to privacy concerns, the extensive time required for detailed labeling by medical experts, and the complexity of surgical actions. Consequently, weak supervision emerges as a pivotal approach, enabling the utilization of limited or imprecise labels to train robust models capable of understanding and recognizing diverse surgical activities. Looking forward, the integration of domain knowledge, such as surgical ontologies and procedural guidelines, into learning frameworks holds the potential to mitigate the limitations posed by weak labels. Additionally, the exploration of unsupervised and semi-supervised methods, combined with weak supervision, could provide new pathways for leveraging unlabelled video data effectively. Collaboration between computer scientists, clinicians, and domain experts is essential to develop more sophisticated algorithms that can understand and predict surgical dynamics accurately.

\noindent \textbf{Few-shot Learning.}
Few-shot learning approaches aim to develop models that can generalize from very limited labeled data, a scenario commonly encountered in the medical field due to the high cost, privacy issues, and time constraints associated with annotating surgical videos. In the context of surgery, these methods are particularly valuable as they allow for the recognition and understanding of surgical actions, tools, and phases from only a handful of examples, thereby facilitating broader applicability across diverse surgical procedures and settings.

\noindent \textbf{Domain Generalization.}
Domain Generalization (DG) techniques are increasingly vital as they allow models to be robust and applicable across different hospitals, surgical procedures, and patient demographics, without the need for retraining. This is particularly crucial in surgical video analysis, where the variance in lighting, surgical techniques, equipment, and individual patient anatomy can vastly differ.

\end{document}